# Checkovid: A COVID-19 misinformation detection system on Twitter using network and content mining perspectives

*Sajad Dadgar, Mehdi Ghatee[*]*

*Department of Mathematics and Computer Science, Amirkabir University of Technology*

*No. 424, Hafez Avenue, Tehran 15875-4413, Iran*

***Abstract***

During the COVID-19 pandemic, social media platforms were ideal for communicating due to social isolation and quarantine. Also, it was the primary source of misinformation dissemination on a large scale, referred to as the infodemic. Therefore, automatic debunking misinformation is a crucial problem. To tackle this problem, we present two COVID-19 related misinformation datasets on Twitter and propose a misinformation detection system comprising network-based and content-based processes based on machine learning algorithms and NLP techniques. In the network-based process, we focus on social properties, network characteristics, and users. On the other hand, we classify misinformation using the content of the tweets directly in the content-based process, which contains text classification models (paragraph-level and sentence-level) and similarity models. The evaluation results on the network-based process show the best results for the artificial neural network model with an F1 score of 88.68%. In the content-based process, our novel similarity models, which obtained an F1 score of 90.26%, show an improvement in the misinformation classification results compared to the network-based models. In addition, in the text classification models, the best result was achieved using the stacking ensemble-learning model by obtaining an F1 score of 95.18%. Furthermore, we test our content-based models on the Constraint@AAAI2021 dataset, and by getting an F1 score of 94.38%, we improve the baseline results. Finally, we develop a fact-checking website called Checkovid that uses each process to detect misinformative and informative claims in the domain of COVID-19 from different perspectives.

## 1. Introduction

As the COVID-19 pandemic broke out, the concerns around misinformation, which can also pose serious problems for public health and societies, have increased. As a primary news source, social media can propagate false and low-quality information and distort trustworthy news about different topics. In the context of the COVID-19 pandemic, incorrect contents disseminate faster and easier than the virus (Munich Security Conference, 2020). So, the pressing need to curb the spread of COVID-19 misinformation has led to increasing active fact-checking organizations worldwide during the pandemic. Due to unprecedented increases in misinformation and public thirst for verified information, fact-checking organization, where the domain experts and journalists analyze data to debunk false and misleading information, fall behind to respond to COVID-19 information immediately. Therefore, to enhance the impact of these organizations, most existing research has proposed various machine learning models to automatically detect false information that generally refers to misinformation, disinformation, and fake news.

[*] Corresponding author: Mehdi Ghatee, Associate Professor of Computer Science, Department of Mathematics and Computer Science, Amirkabir University of Technology, Tehran, Iran.
Email address: s.dadgar@aut.ac.ir (S. Dadgar), ghatee@aut.ac.ir (M. Ghatee)

Generally, we avoid using the "Fake News" term and focus on misinformation that is false information, regardless of whether it disseminates intentionally or not. We start our work by constructing two datasets that contain misinformative and informative tweets of COVID-19 collected from fact-checking websites and reliable organizations. Then, we introduce a misinformation detection system by network-based and content-based processes, which are developed to classify the veracity of the tweets from disparate aspects. In each process, we use Natural Language Processing (NLP) techniques and supervised machine learning algorithms divided into traditional machine learning, ensemble learning, and deep learning algorithms. In the network-based models, we detect misinformation based on tweets and users features extracted from the Twitter API and linguistic features hidden in the tweets' content. In the content-based process, we develop two types of models that detect misinformation directly from tweets' content: text classification models and similarity models. We train different text classification models in paragraph-level and sentence-level. Also, we introduce novel similarity models to detect misinformation based on the similarity between tweets. In the last step, we propose a COVID-19 fact-checking website called Checkovid that includes proposed processes and can automatically detect misinformation.

The rest of the paper is organized as follows. In section 2, we discuss prior works related to COVID-19 and Coronavirus. In section 3, we describe our methodology and discuss the structure of our proposed system. Then, Section 4 discusses the experiments, case study, hyperparameter tuning process, presents the results, and evaluates the performance of models. Finally, we conclude the paper by summarizing our contributions and discuss future work in section 5.

## 2. Related works

In this section, we review the previous works in the context of COVID-19 and Coronavirus. Due to the proliferation of large volumes of false content during the pandemic, the study around COVID-19 related misinformation became a popular area of research. Thus, many studies have discussed the impact and the various characteristic of misinformation. Also, many researchers have proposed various automated misinformation and fake news detection models on different datasets.

### *2.1. Datasets*

In order to avoid the spread of COVID-19 related false contents, several datasets have been released that are specifically concerned with the misinformation and fake news in this domain that could help researchers develop different machine learning-based models. See, e.g.,

- ReCOVery(Zhou, Mulay, Ferrara, & Zafarani, 2020), contains the news content and related multimodal information;
- CoAID (Cui & Lee, 2020), A healthcare misinformation dataset contains fake news on websites and social media;
- CMU-MisCOV19 (Memon & Carley, 2020), A English annotated tweets dataset that includes misinformed, informed, and irrelevant users;
- A misinformation dataset on Twitter contains fact-checked claims collected from fact-checking websites (Shahi, Dirkson, & Majchrzak, 2021).

In terms of multilingual datasets, examples include,

- FakeCovid (Kishore Shahi & Nandini, 2020), a fact-checked dataset in 40 languages collected from fact-checking websites;
- MM-COVID (Y. Li, Jiang, Shu, & Liu, 2020), a multilingual and multidimensional dataset, including fake news content, social engagement, and spatial-temporal information;

- An Arabic and English Twitter misleading information dataset about COVID-19 proposed by Elhadad, Li, and Gebali (2021);

Also, from datasets with language other than English,

- ArCOV19 (Haouari, Hasanain, Suwaileh, & Elsayed, 2020), contains Arabic misinformation tweets;
- A Chinese fake news dataset containing multimedia information (C. Yang, Zhou, & Zafarani, 2021).

## 2.2. *detection models*

Machine learning algorithms consist of applying mathematical and statistical methods that enable models to learn on datasets. Therefore, researchers step forward and employ different natural language processing and feature extraction techniques to develop machine learning-based models for identifying misinformation. Elhadad, Li, and Gebali (2020) assessed ten supervised machine learning algorithms with seven feature extraction techniques to detect COVID-19 misleading textual information. Logistic Regression, Decision Tree, and Neural Network provide the best results. Al-Rakhami and Al-Amri (2020) proposed a stacking-based ensemble learning model by integrating six machine learning algorithms to handle Twitter misinformation detection using tweet-level and user-level features. The proposed ensemble learning model performs better than other single machine-learning-based models. Hossain (2020) released a dataset containing COVID-19 misconceptions and their misinformative and informative expressions on Twitter and assessed the performance of misinformation detection systems on other misinformation relating to COVID-19. Additionally, a text similarity model has been proposed that can detect whether or not a post is relevant to the COVID-19 related misconception.

As a result of limited labeled datasets, most studies developed semi-supervised machine learning models for misinformation classification. For instance, Paka, Bansal, Kaushik, Sengupta, and Chakraborty (2021) proposed Cross-SEAN, semi-supervised neural attention to detect fake news and showed that this model outperformed seven state-of-the-art models. Also, many researchers suggested transformer-based models that are composed of unsupervised pre-training and supervised fine-tuning. Studies have demonstrated transformer-based models achieved better performance than supervised machine learning models in text classification. For example, Pranesha, Farokhenajdb, Shekhara, and Vargas-Solarc proposed CMTA, a multilingual BERT model, trained on multiple languages to classify multilingual tweets into three classes. Serrano, Papakyriakopoulos, and Hegelich (2020) proposed a multi-label classifier using transfer learning to detect COVID-19 misinformation videos on YouTube based on user comments and showed that misinformative videos contain a higher amount of conspiratorial comments. Kumari, Ashok, Ghosal, and Ekbal (2021) proposed a multitask learning misinformation detection framework. The result revealed the improvement in the fake news detection task with two auxiliary tasks: novelty detection and emotion detection. Ayoub, Yang, and Zhou (2021) proposed a prediction model with DistilBERT and SHAP. Results indicate the high performance of the DistilBERT model to detect misinformation about COVID-19. Hamid et al. (2020) proposed two independent approaches: content-based and structure-based for fake news detection. The content-based task relied on Bag of Words and BERT embedding, and the structure-based relied on Graph Neural Network and revealed that the best results are obtained with Bag of Words. Kar, Bhardwaj, Samanta, and Azad (2020) proposed a multilingual BERT embedding model to detect fake news about COVID-19 from Twitter, trained on multiple Indic-Languages fake news datasets. Since it is helpful to detect false content as soon as possible, early-stage detection models have been proposed, such as ENDEMIC (Bansal, Paka, Nidhi, Sengupta, & Chakraborty, 2021), a semi-supervised co-attention network for early detection of COVID-19 fake news on a developed dataset called ECTF, and the results indicated the outperforming of the ENDEMIC compared to nine state-of-the-art models in early-stage fake news detection. Propagation2Vec (Silva, Han, Luo, Karunasekera, & Leckie, 2021) is a network-based framework for the early detection of fake news. Based on the obtained result, Propagation2Vec performs better than state-of-the-art fake news detection models while having access to the early stage propagation networks.

## 2.3. *Constraint@AAAI2021 shared task*

In addition, several of these supervised machine learning and transformer-based models contributed to the ConstraintAAAI@2021 shared task using the dataset mentioned in (Patwa et al., 2020) for fighting an infodemic. For example, Felber (2021) presented their contribution to the task by applying supervised traditional machine learning algorithms using several linguistic features on the dataset and achieved the best performance with the SVM model. Wani, Joshi, Khandve, Wagh, and Joshi (2021) and Gundapu and Mamid (2021) compared the performance of various supervised machine learning, deep learning, and transformer-based models using several evaluation metrics. The results revealed that transformer-based models are better than other models. Furthermore, Das, Basak, and Dutta (2021) proposed an ensemble model consisting of a pre-trained model, and the results improved by using an ensemble mechanism with Soft-voting and achieved better results with a heuristic post-processing technique. Finally, Glazkova, Glazkov, & Trifonov (2020) proposed a fake news detection based on CT-BERT (COVID-Twitter-BERT) and ensemble learning and achieved the best result among other models (first place in the ranking) in the Constraint@AAAI2021 shared task.

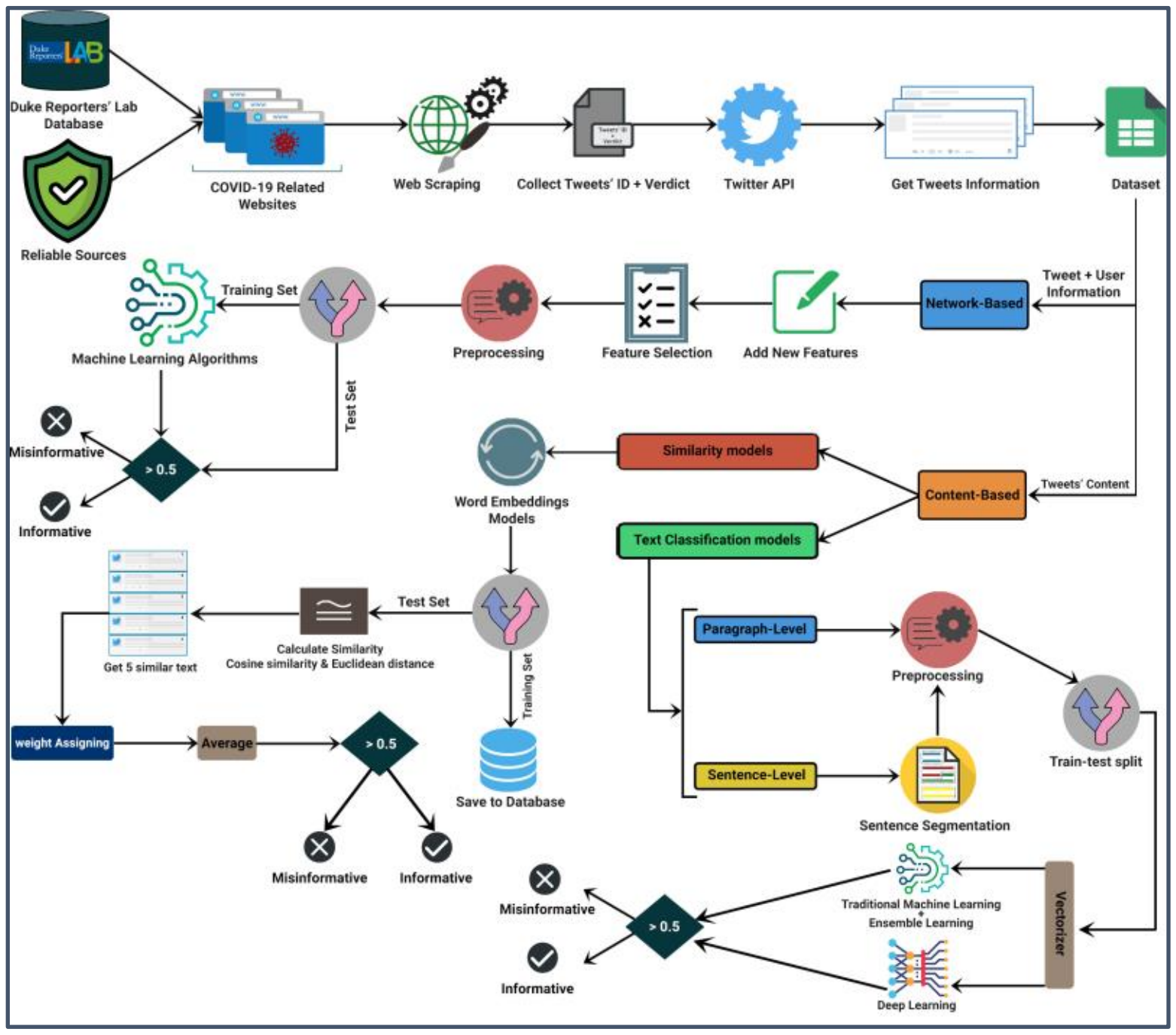

*Figure 1. Structure of the proposed misinformation detection system*

## 3. Methodology

This paper aims to focus on different methods and techniques to build a misinformation detection system. Figure 1 illustrates the structure of our proposed misinformation detection system. Studies showed that determining the veracity of a claim without further fact-checking is challenging, and humans have achieved an average of 54% accuracy in the task of distinguishing between a lie and truth (Bond & DePaulo, 2006). In a crisis (e.g., COVID-19 pandemic), determining false content can be more challenging because of extensive interaction between users and the fast dissemination of related content in social media. Therefore, identifying misinformation has shifted to automated methods in the last few years. This paper introduces a COVID-19 misinformation system that contains two automated processes to address this issue: network-based and content-based, which includes text classification models and similarity models. The network-based models classify a tweet based on the social characteristics of the tweet and the user who written it. Classifiers of the content-based process distinguish misinformation by extracting features from texts and recognize the similarity of the tweet to other labeled tweets. Below we will discuss each process and explain the models in detail. The data used in this paper and the implementation of all models is available through GitHub[1].

[1] https://github.com/sajaddadgar/A-COVID-19-misinformation-detection-system-on-Twitter-using-network-content-mining-perspective.git

*Table 1. Dataset example*

| | |
|---|---|
| **Tweet:** | "Shanghai Government Officially Recommends Vitamin C for COVID-19" |
| **Verdict:** | Misinformative |
| **Tweet:** | "BILL GATES EXPLAINS THAT THE COVID VACCINE WILL USE EXPERIMENTAL TECHNOLOGY AND PERMANENTLY ALTER YOUR DNA" |
| **Verdict:** | Misinformative |
| **Tweet:** | "The coronavirus was engineered by scientists in a lab using well documented genetic engineering vectors that leave behind a fingerprint." |
| **Verdict:** | Misinformative |
| **Tweet:** | "CDC recommends men shave their beards to protect against coronavirus." |
| **Verdict:** | Misinformative |
| **Tweet:** | "COVID-19 virus can be transmitted in areas with hot and humid climates." |
| **Verdict:** | Informative |
| **Tweet:** | "Taking a hot bath does not prevent the new coronavirus disease." |
| **Verdict:** | Informative |
| **Tweet:** | "Drinking alcohol DOES NOT protect you against COVID-19 and can be dangerous." |
| **Verdict:** | Informative |
| **Tweet:** | "The coronavirus disease (COVID-19) is caused by a virus. NOT by bacteria." |
| **Verdict:** | Informative |

## 3.1. *Data Collection*

In this paper, we construct two datasets. The first dataset consists of informative and misinformative tweets collected from fact-checking websites and reliable organizations that publish trustworthy information. The second dataset consists of informative and misinformative sentences derived from the first dataset.

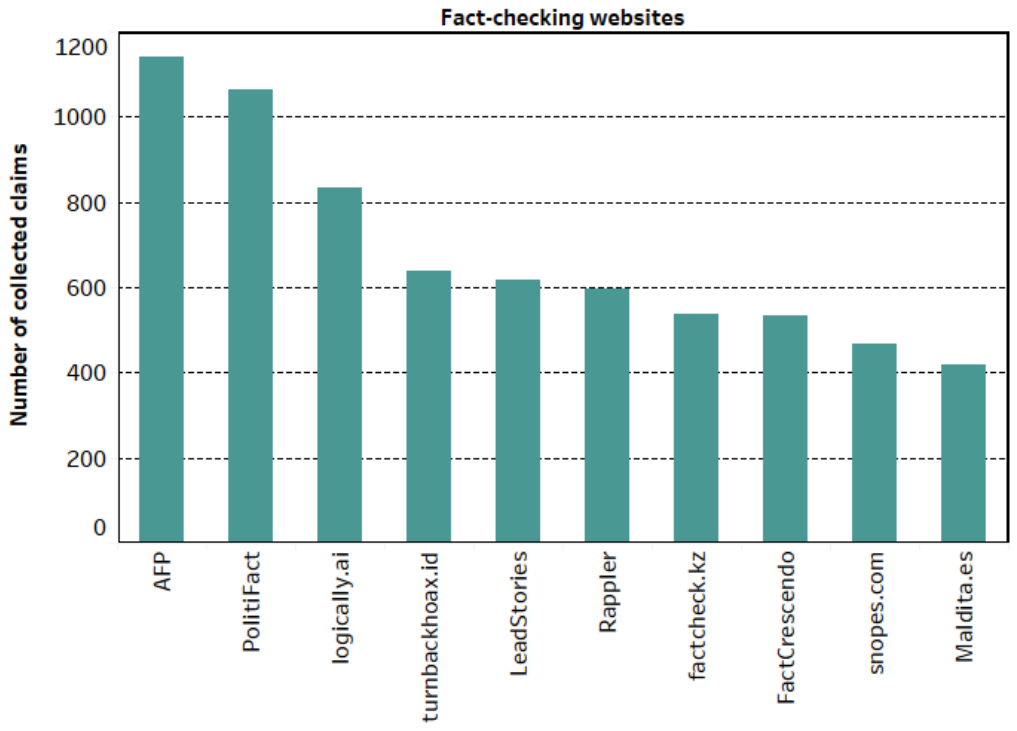

*Figure 2. Distribution of Top-10 Fact-checking website*

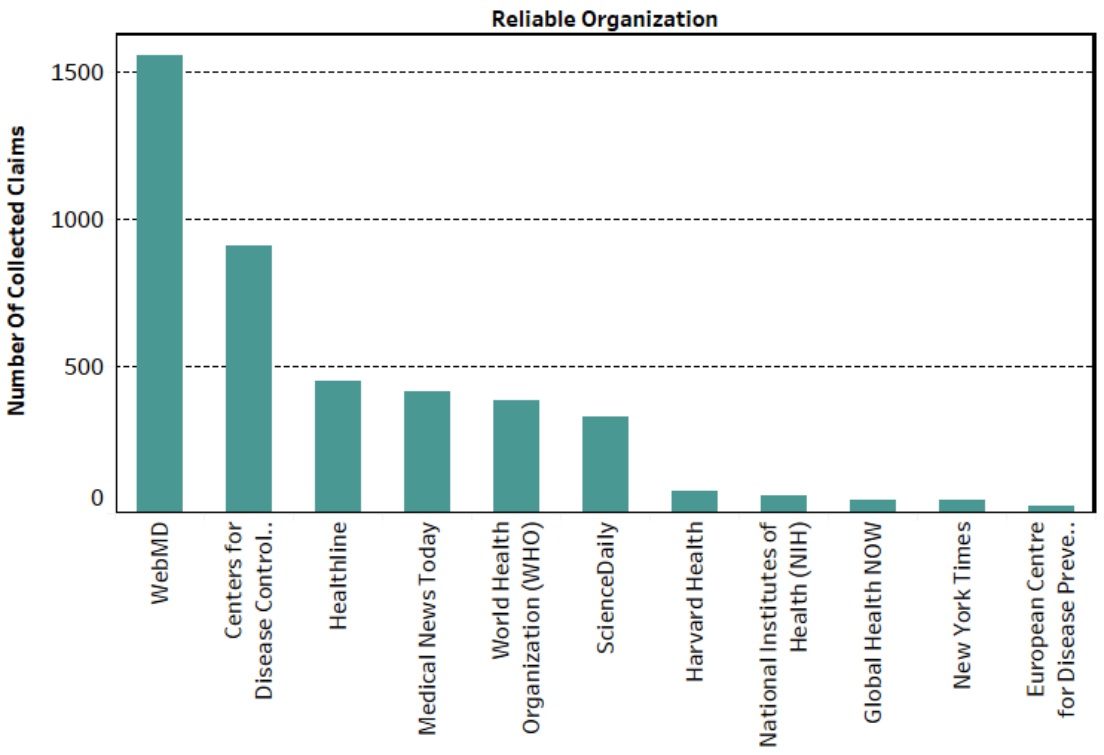

*Figure 3. Distribution of Reliable organizations*

### 3.1.1. *Dataset I*

As shown in Figure 1, we first need to collect COVID-19 related tweets with reliable labels that indicate their veracity. Therefore, for our study, we gather data from two primary sources. First, we collect claims about COVID-19 from fact-checking websites that have taken on the mission of fact-checking rumors, particularly health claims and political claims. Second, reliable organizations that consist of healthcare/public-health organizations with expertise relating to the SARS-COV-2 virus and reliable media that inform people with facts and reliable information during the pandemic (Figure 3). To get all fact-checking websites, we use the database maintained by the Reporters' lab at Duke University[2] that contains the list of global fact-checking sites. We were able to collect 153 fact-checking sites that covered COVID-19 and Coronavirus news. We plot the 10 fact-checking websites that covered the most covid-19 related claims in 2020 in Figure 2. We take the same automated approach mentioned in (Kishore Shahi & Nandini, 2020; Shahi et al., 2021) to retrieve tweets from these websites. We use Beautifulsoup, which is a python library to crawl and parsing HTML documents. We crawl HTML content referred to tweets and look for all anchor elements (or <a>) that are created a hyperlink to tweet's URL in their HREF attribute. In other words, the anchor elements that their HREF attribute is in the form of https://twitter.com/username/status/tweet_id. We fetch tweet IDs and save them with the rating of the fact-checking website that verified them. We do not consider IDs mentioned as sources in these websites because they verify the claim, and their verdict is unclear. Also, because each of these fact-checking websites uses different ratings for the veracity of claims, we divide them based on their rating into informative and misinformative classes and exclude all of the tweets' IDs with other ratings, such as unknown, mixture, and ambiguous. Moreover, since the ratio of the data with misinformative labels is more than informative labels and our dataset will be imbalanced, we also crawl the public health websites and other reliable organizations and collect tweet IDs to add more informative data. Studies indicated that the Twitter account of these organizations had the lowest rate of unverifiable information among other accounts (Kouzy et al., 2020). Finally, we extract tweet information and user engagement via Twitter API from the obtained tweet IDs. After removing tweets with languages other than English, we end up with 9012 misinformative and 9133 informative COVID-19 related tweets written by 15060 unique users. Some examples of misinformative and informative tweets are illustrated in Table 1. This dataset is collected from 01-09-2020 to 28-12-2020 and contains tweets from December 2019 until December 2020. The timeline of the collected tweets is shown in Figure 4 and indicates the fast propagation of the misinformative tweets than informative tweets during the first few months of the pandemic.

[2] https://reporterslab.org/fact-checking/

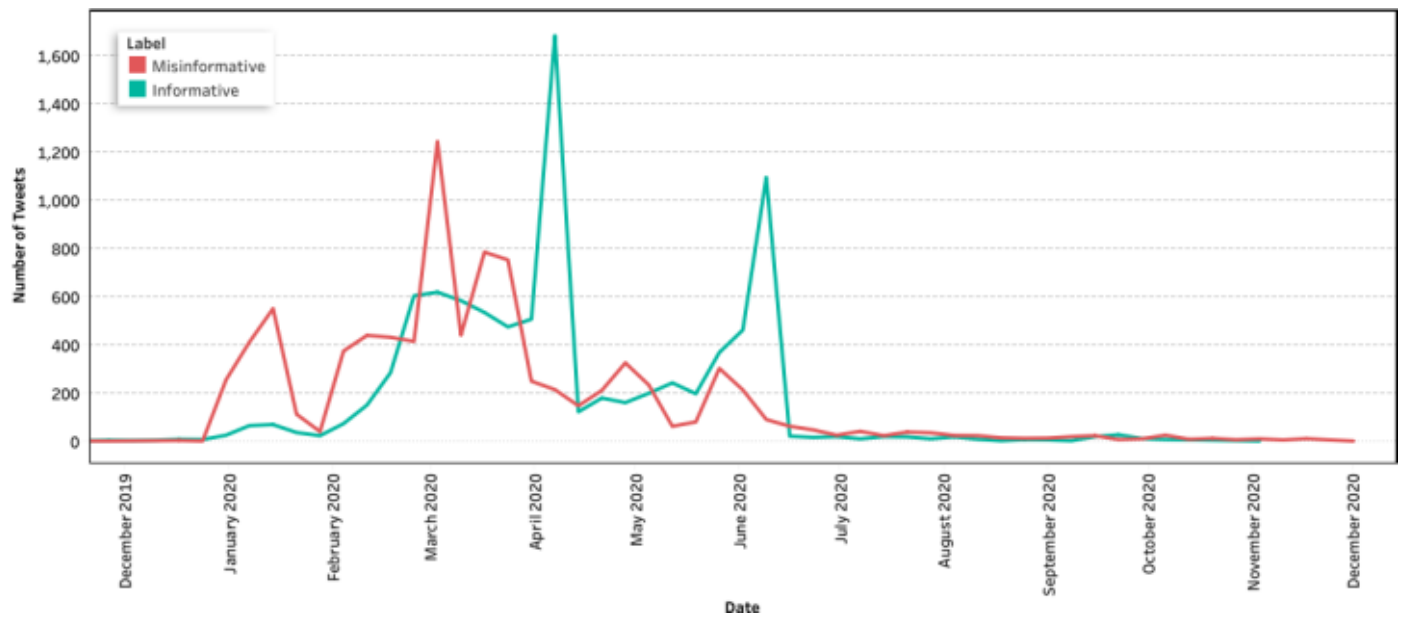

*Figure 4. Timeline of misinformative and informative tweets created from December 2019 to December 2020 in Dataset I*

### 3.1.2. *Dataset II*

To detect misinformation at the sentence level, we use the content of the tweets in D*ataset I* and segment them into sentences. Sentence segmentation is the process of identifying sentences among groups of words. To perform sentence segmentation with high accuracy, we use Spacy, a python library designed for advanced NLP. We use this library instead of splitting sentences using the period mark ('.') because the period has been used for different purposes, such as abbreviations and numbers, so segmenting sentences by the period mark is not practical. After segmenting the sentences, we assign each sentence to the verdicts in D*ataset I.* Since there are many meaningless sentences, we exclude them and develop 15635 sentences with informative and misinformative labels.

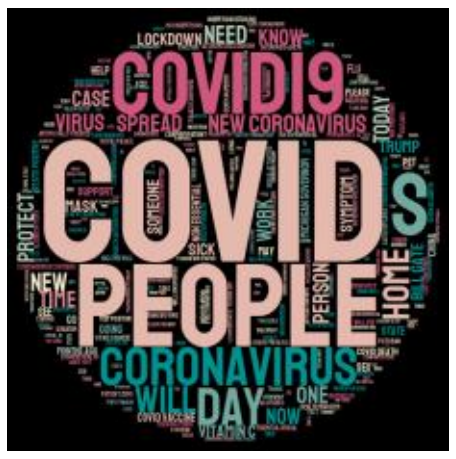

*Figure 5. Informative tweets in Dataset I*

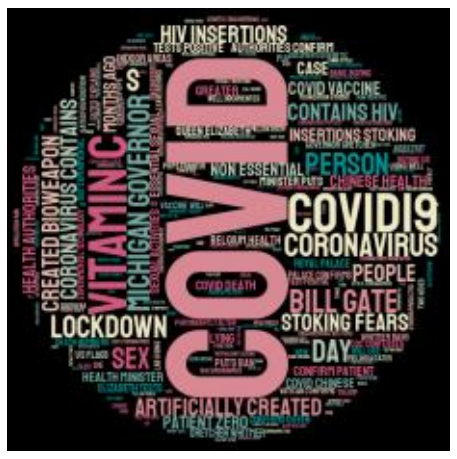

*Figure 6. Misinformative tweets in Dataset I*

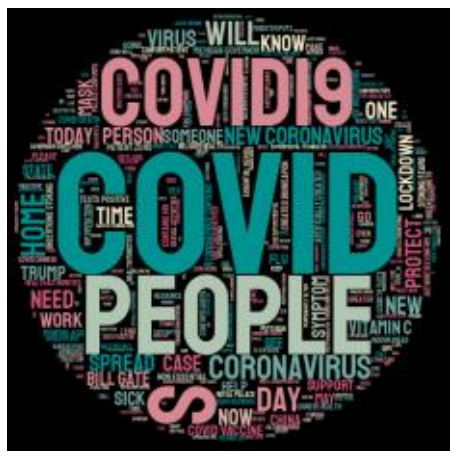

*Figure 7. All tweets in Dataset I*

## *3.2. Network-Based*

Social media platforms not only allow people to engage with each other and exchange information in extensive and beneficial ways but also provide immense opportunities for misinformed users to disseminate misinformation actively.

Studies showed that these user engagements and other network properties, such as post and user information, could serve as features that have the potential for distinguishing misinformation with high accuracy in the limited domain (Conroy, Rubin, & Chen, 2015). In this process, we utilize different network and linguistic features related to tweets, users, and the content of tweets and fed them into supervised machine learning models to train and predict misinformation from a network perspective. D*ataset I* contain 15 tweet features and 14 features about the user who written it. To develop more robust models and achieve better performance, we create additional features in the domain of COVID-19 and leverage some features mentioned in (S. Li; Pérez-Rosas, Kleinberg, Lefevre, & Mihalcea, 2017) and categorize these features into tweet features and user features. The description of each category of features are as follow:

- **Tweet Features:**

  Social media users express their opinions and emotions toward misinformation and respond to them in distinctive ways. Thus, extracting tweet features can provide some patterns and clues that help to distinguish misinformation. These features focus on identifying beneficial social characteristics and linguistic features of tweets, which are latent in the content of the tweets. These features include tweets engagement that the Twitter API provided, such as the count of retweets and likes. On the other hand, linguistic features extracted from the content of tweets consist of readability scores, sentiment, syntactic features such as parts-of-speech (POS) tagging and punctuations, and lexical features in both character-level and word-level as the number of capital characters, frequency of stopwords.

- **User Features:**

  Tweets can be created and spread by various user accounts, such as people, organizations, bots, and other non-human accounts. Thus, capturing user engagement and their Twitter account profile information can provide helpful information to identify misinformative from informative tweets. These features include different user characteristics such as account creation date, number of followings and followers, number of tweets that the user has been published.

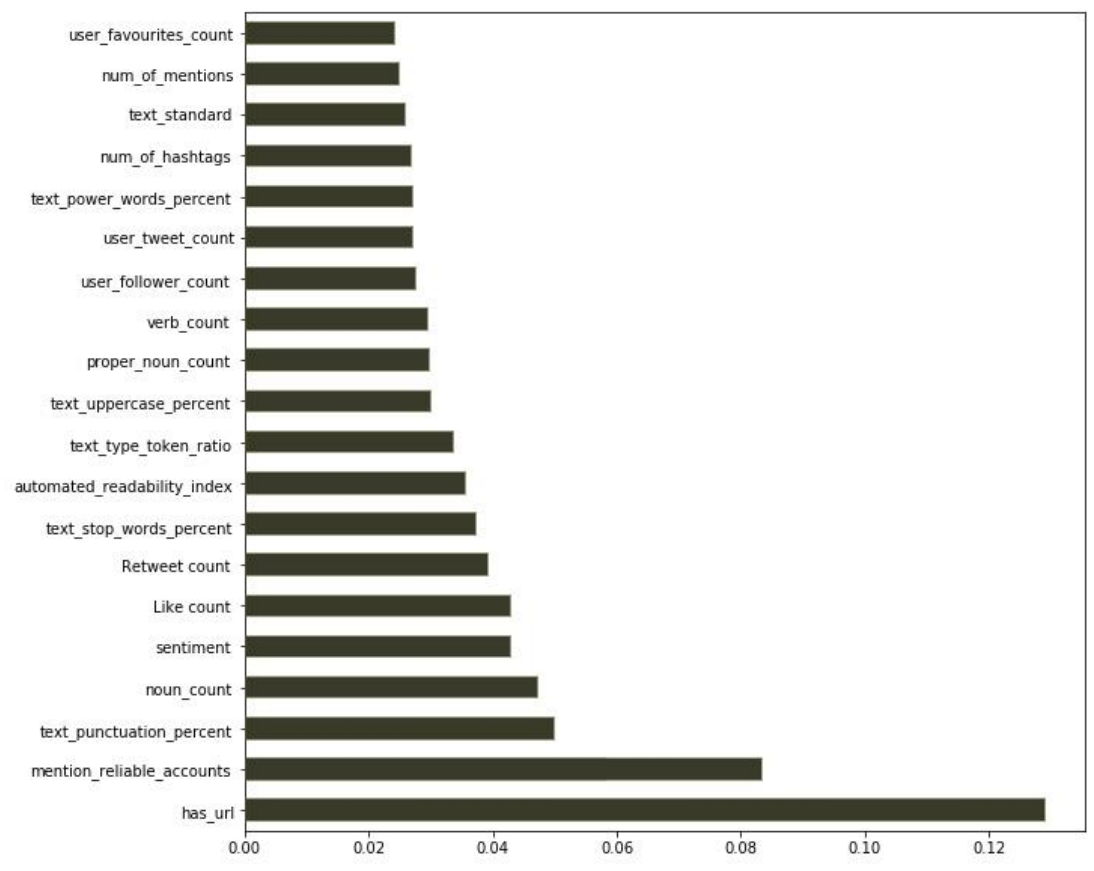

*Figure 8. Feature importance*

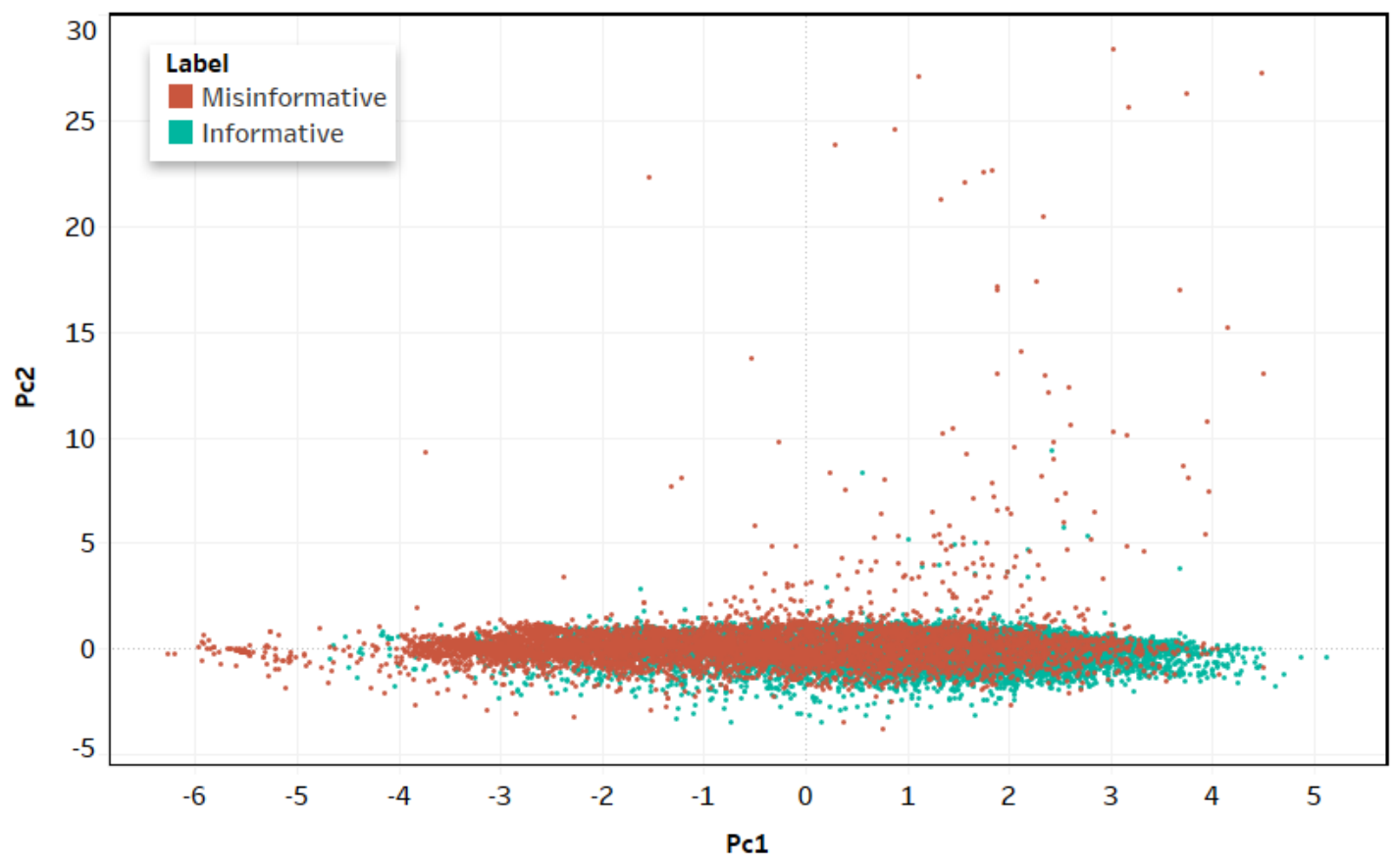

*Figure 9. Data visualization (PCA with two dimensions)*

### 3.2.1. *Feature Selection*

The features used to train on machine learning models significantly influence the performance. Therefore, firstly, we prepare them for model training. Machine learning algorithms require that input variables be numerical. Consequently, we use one-hot encoding to convert the categorical features into numbers. Also, since the range of the features is varied, we use standardization so that each feature contributes approximately proportional to the model training. Second, because some of these features are irrelevant and redundant attributes that do not effectively contribute to the model training, it is desirable to eliminate these features to reduce the computational cost and improve the accuracy of our models. Therefore, we perform Recursive Feature Elimination (RFE), a wrapper-type feature selection algorithm that works by recursively eliminating features, constructing a model on the remaining features, and ranking the importance of the features according to the evaluation results of the model. Overall, we extract 43 features from both tweet and user features. We use a random forest classifier as the core model and select the top 20 features shown in Figure 8. In Table 2, we provide a list of all the extracted features and their descriptions. The bold features indicate that the features are chosen and used in the final models. Also, Figure 9 represents the data in two dimensions using Principal Component Analysis (PCA). In the next step, we classify misinformation by training different machine learning algorithms based on these features.

*Table 2. Description of extracted features*

| Feature name | Description | Comment |
|---|---|---|
| Tweet features | | |
| Tweet_date | The date that the tweet was created in milliseconds | Extracted from Twitter API. |
| tweet_type | The type of tweet (tweet, retweet, quote, reply) | |
| **Like count** | Number of likes | |
| **Retweet count** | Number of retweets | |
| Possibly sensitive | Whether the URL in the tweet contains content identified as sensitive content | |
| **sentiment** | The sentiment score of the tweet content in a range of [-1, 1] | To determine the tweet's sentiment, we used TextBlob, a python library to process textual data. |
| **mention_reliable_accounts** | Whether the tweet mentions a reliable Twitter account | Reliable accounts obtained from a list created by Jonathan Oppenheim consist of Twitter accounts of scientists, journalists, and [3]organizations with expertise relating to the SARS-CoV-2 pandemic. |
| **has_url** | Whether the tweet contains a URL | Extracted from the content of the tweet. |
| **num_of_mentions** | Number of mentions | |
| **num_of_hashtags** | Number of hashtags | |
| emoji_count | Number of emojis | |
| **text_uppercase_percent** | The percentage of capital letters | |
| **text_punctuation_percent** | The percentage of punctuation marks | |
| **text_stop_words_percent** | The percentage of stop words | |
| **verb_count** | Number of verbs | We used the Part-of-speech tagger (POS tagger) using NLTK and kept a count of each tag's appearance in the text. |
| **proper_noun_count** | Number of proper nouns | |
| **noun_count** | Number of nouns | |
| pronoun_count | Number of pronouns | |
| adjective_count | Number of adjectives | |
| **text_power_words_percent** | The percentage of power words | Extracted features with the same approach (S. Li) and with the same words. |
| text_casual_words_percent | The percentage of casual words | |
| text_tentative_words_percent | The percentage of tentative words | |
| text_emotion_words_percent | The percentage of emotional words | |
| text_swear_words_percent | The percentage of swear words | |
| **text_type_token_ratio** | The Type Token Ratio (TTR) | We used lexicalrichness, a python library to compute textual lexical richness measures. |
| flesch_reading_ease | The Flesch Reading Ease score | We used Textstat, a python library to calculate statistics from text, such as readability, complexity, and grade level. |
| smog_index | The SMOG grade | |
| flesch_kincaid_grade | The Flesch-Kincaid Grade Level | |
| **automated_readability_index** | The Automated Readability Index (ARI) | |
| dale_chall_readability_score | The Dale-Chall readability score | |
| linsear_write_formula | The Linsear Write Formula | |
| gunning_fog | The FOG index | |
| **text_standard** | Determines the estimated school grade based on readability tests | |
| difficult_words | Number of difficult words | |
| User features | | |
| user_created_at | The date that the user account was created in milliseconds | Extracted from Twitter API |
| **user_follower_count** | Number of the user's followers | |
| user_following_count | Number of the user's followings | |
| **user_favourites_count** | Number of the tweets that the user has liked | |
| user_verified | Whether the user has verified Twitter account | |
| **user_tweet_count** | Number of the tweets that the user has published | |
| has_user_url | Whether the user has a URL on its profile | |
| user_geo | Whether the user has a location | |
| user_profile | Whether the user has a profile image | |

[3] https://www.ucl.ac.uk/oppenheim/Covid-19_tweeps.shtml

### 3.2.2. *Model Construction*

After preprocessing data and selecting the best discriminating features, we perform seven classification models to detect misinformation using supervised machine learning algorithms, divided into traditional machine learning, ensemble learning, and deep learning models. We implement Logistic Regression, Naïve Bayes, Support Vector Machine (SVM), Decision Tree as traditional machine learning models. In ensemble learning models, multiple learning models are combined to make more robust models. For ensemble learning, we use Random Forest and Stacking (Stacked Generalization) algorithms. The Random Forest model comprises 100 independent decision tree classifiers trained on different sub-samples of the training set. On the other hand, the stacking model consists of the stacking output of the individual base-learners on the training set and makes the final prediction with a meta-learner. We use Naïve Bayes, SVM, and Decision three as base-learners and Logistic Regression as meta-learner. As a deep learning model, we develop an Artificial Neural Network (ANN). To achieve the best result, we tune the hyperparameters for each model that will be discussed in Section 4.3.1.

## 3.3. *Content-Based*

In addition to the network-based process, which uses features derived from user engagements and tweets information, we can use machine learning models and NLP techniques to detect misinformation from the tweets' content directly. In this process, we use two types of models: text classification models and similarity models.

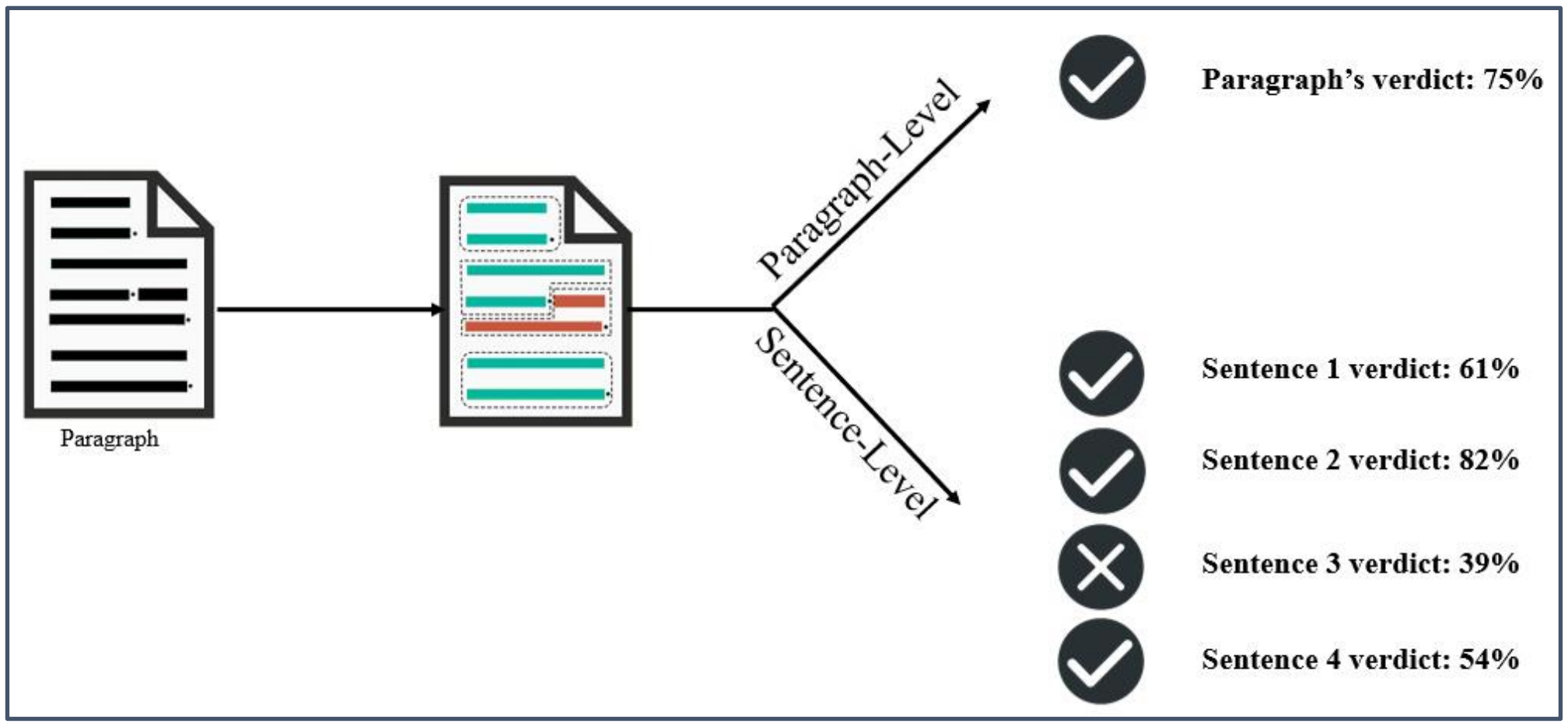

*Figure 10. Paragraph-level vs. Sentence-level*

### 3.3.1. *Text classification models*

Misinformative texts express distinctive patterns that machine learning models can identify. In essence, text classification models can capture discriminative patterns from the linguistic and style of the tweet's content. Therefore, researchers have begun to address the issue of combat misinformation and have developed various machine learning-based models and NLP techniques duo to identify these distinctive patterns in the news contents and claims about different topics (e.g., healthcare, politics). However, most misinformative users use various strategies to avoid getting caught (Conroy et al., 2015). One way to deceive people into believing false content is to hide the misinformation among several facts and informative claims. The chance of distinguishing misinformation decreases when it combines with some informative sentences in a paragraph. Therefore, these models aim to identify tweets with misinformation in two levels: paragraph-level, sentence-level. Figure 10 illustrates the structure of each level for classifying misinformation. For both paragraph and sentence levels, we do the same steps to develop the text classification models, except that we perform these models on the different datasets for each level. At the paragraph-level, we use the content of the tweets in *Dataset I*, and at the sentence-level, we use *Dataset II*. First, before getting the machine learning models to learn from our corpus, we do the preprocessing step using NLP techniques and convert the textual data into a more digestible form to improve the performance of the models.

### *3.3.1.1. Preprocessing*

The following steps of preprocessing are used in the text classification models:

- **Remove hyperlinks:** There are many hyperlinks in tweets, but they do not add any value to textual data. Therefore, we get rid of these hyperlinks.
- **Handling special characters:** Special characters are neither numbers nor alphabets, but users use them in their text. Therefore, we handle them depending on their utilization in the text. For example, we remove "$" and "#" symbols, but we keep their word and number because users may use them in their sentences and write a word with hashtags. Moreover, we convert "&" into "and", remove "@", which is used for mentioning other users, and "RT", which some users use in their tweets, and indicate whether a tweet is a repost of others content.
- **Lowercase conversion:** We convert the entire text in our corpus into lowercase due to eliminating sensitivity to uppercase and lowercase letters in our models.
- **Stemming:** For grammatical reasons, people use different derivative forms of a similar-meaning word in their text. Therefore, we use the Porter stemming algorithm to reduce words to their root word.
- **Remove punctuations and stopwords:** We also remove punctuations and stopwords from the texts. Punctuations contain marks that help the structure of a sentence. On the other hand, stopwords are frequent words used in texts that do not add much value to the meaning of a sentence. In addition, we customize the stopwords and keep negation words (e.g., no, not) in our corpus and remove other stopwords. Moreover, because "who" is a stopword and also it is a common word in the domain of COVID-19 pandemic duo to many discussions around the World Health Organization (WHO), before lowercasing the text, we convert some forms of this word (e.g., WHO, "WHO", "Who", "who") into "world health organization". Also, we only consider this conversion when "WHO" is in the middle of the text or if it is at the beginning of the text, it does not end with a question mark. With customization of the stopwords, we improved the accuracy of the following models by 0.86 on average.

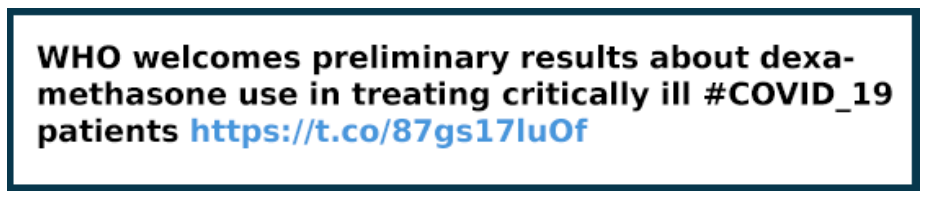

*Figure 11. Raw text*

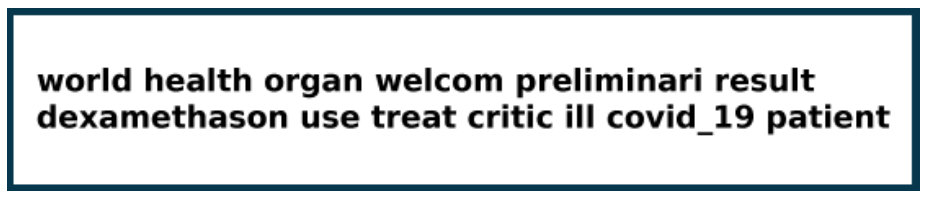

*Figure 12. Preprocessed text*

### *3.3.1.2. Model Construction*

In this section, we train machine learning models categorized into traditional machine learning, ensemble learning, and deep learning models on the content of tweets to predict the veracity of the tweets. Since these models require taking vectors as input, for each model, we use feature extraction techniques to convert the preprocessed text data into vectors. Moreover, we tune each model and train them with the optimum values of hyperparameters to develop the models with the highest accuracy. We will discuss the hyperparameter tuning process and the results of models in Section 4.3.1.

- **Traditional Machine Learning Models:**

For extracting features from text data, various techniques have been used, such as Bag of Words (BOW) and Term Frequency-Inverse Document Frequency (TF-IDF). Although Bag of Words works fine in different NLP tasks, it is less practical since it suffers from some shortcomings, such as ignoring word order, considering each word with equal importance, and sparsity. Therefore, we use TF-IDF which consists of multiplying two metrics to extract features from textual data:

- Term Frequency (TF) measures the importance of a word in a document.
- Inverse Document Frequency (IDF) measures how common or rare a word is in all documents.

Thus, we vectorize all of the tweets and convert them into a matrix of TF-IDF features. The extracted features are fed into Logistic Regression, Naïve Bayes, Support Vector Machine (SVM), and Decision Tree model.

- **Ensemble Learning Models:**

  To improve the results of the traditional machine learning models, we develop Random Forest and Stacking as ensemble learning models based on the traditional machine learning models with the same structure in the network-based process.

- **Deep Learning Models:**

  With the ability of deep learning models to learn high-level features, existing computational capabilities on a large amount of data, we implement four supervised deep learning models, including Long Short-Term Memory (LSTM), Bidirectional LSTM (Bi-LSTM), Convolutional Neural Network (CNN), and CNN+LSTM models. Before implementing these models, we prepare the preprocessed data by encoding each word in the corpus into a unique integer number. Furthermore, since deep learning models take inputs of the same length and dimension, input texts are padded to the maximum length.

### 3.3.2. *Similarity Model*

Due to the large numbers of active users in social media that exchange information in an interconnected way, a massive flood of information, along with misinformation, is disseminated in a fraction of a second. To amplify misinformation and make a false claim viral, potentially coordinated groups of accounts may work together to post similar false content (K.-C. Yang, Torres-Lugo, & Menczer, 2020). As a result, there are numerous similar contents on social media platforms with the same veracity. Figure 13 shows an example of similar misinformative claims that are disseminated on Twitter. This section introduces the similarity models that classify the veracity of an unlabeled tweet based on its similarity to labeled tweets in the dataset. We use Cosine similarity (Eq. 1) and Euclidean distance (Eq. 2) as two metrics for measuring the similarity between tweets' content.

$$similarity(a,b) = cos(\theta) = \frac{\vec{a}.\vec{b}}{\|a\|\|b\|} \quad \text{(Eq. 1)}$$

$$d(a,b) = \sqrt{(a_1 - b_1)^2 + (a_2 - b_2)^2} \quad \text{(Eq. 2)}$$

Cosine similarity is a similarity metric that determines whether or not two vectors are similar by measuring the cosine of the angle between two vectors. On the other hand, Euclidean distance measures the shortest distance between two points. To work with these metrics and calculate the similarity between texts in our corpus, we need to convert each tweets' content into a vector of real numbers. Thus, we use word embedding to assign each unique word in the texts into a corresponding vector so that the same-meaning words and words that share common contexts are positioned close to each other in the space. The vector of each tweet is derived by summing up the embeddings of all words in its content. Different types of word embedding models can be used to obtain the vectorized representation of text. We apply the following word embedding models:

- ***Word2vec:*** is a predictive model that uses a two-layer neural network to learn words. There are two model architectures for Word2vec to generate word embeddings: Continuous Bag of Words (CBOW) and Continuous Skip-Gram (Skip-Gram) methods. The objective of the CBOW model is to predict the center word given some context words. However, the Skip-Gram model does reverse the CBOW model and predict the word surrounding given an input word (Mikolov, Sutskever, Chen, Corrado, & Dean, 2013).
- ***Global Vectors (Glove):*** is an unsupervised learning model for word representation learned by constructing a co-occurrence matrix of each word and reducing the matrix's dimension using matrix factorization methods (Pennington, Socher, & Manning, 2014).
- ***FastText:*** A library, a modified version of Word2vec and based on the Skip-gram model. It enables the model to support out-of-vocabulary (OOV) words. Also, word embedding vectors can be averaged to make vector representations of sentences (Bojanowski, Grave, Joulin, & Mikolov, 2017).
- ***COVID-19 Concept Embeddings:*** A pre-trained COVID-19 related word embedding (Newman-Griffis, Lai, & Fosler-Lussier, 2018) trained on the COVID-19 Open Research Dataset (Lu Wang et al., 2020).

To determine whether an unlabeled tweet is misinformative or informative, we convert its content to a vector using a word embedding model and calculate the similarity between the unlabeled tweet and all labeled tweets.

Afterward, we select K texts with the highest cosine similarity score or lowest Euclidean distance to the unlabeled tweet and assign a weight for each similar text. So that a more similar tweet has more weight value and, as a result, has a more effect on classifying the label of the target tweet. Finally, we determine the veracity of the target tweet by computing the weighted average of the K similar texts. In the following, we will discuss how to evaluate these models, how to select the best number for K, and the result obtained for each metrics and word embedding model.

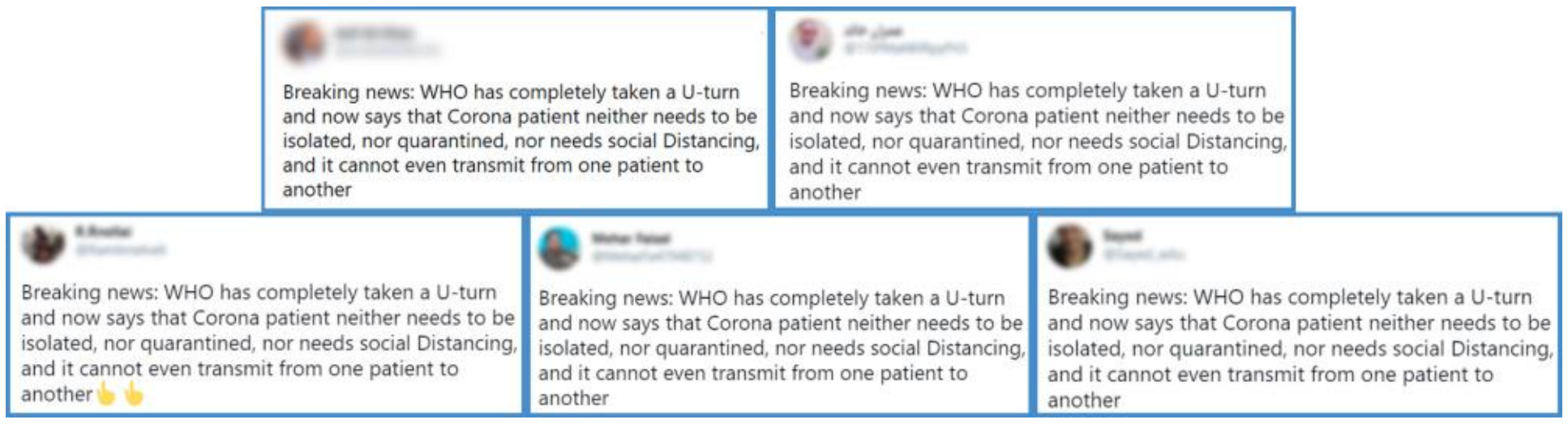

*Figure 13. An example of similar misinformation on Twitter*

## 4. Experiments and Results

In this section, we explain our proposed website for detecting COVID-19 related misinformation as a case study. Also, this section discusses the experimental evaluation and the hyperparameter tuning process of each model. Also, we describe the architecture of the machine learning and deep learning models and compare the obtained results.

### *4.1. Case Study*

We propose an English COVID-19 fact-checking website called Checkovid[4]. To develop Checkovid, we use Django, a python framework for back-end development, along with HTML, CSS, and JavaScript to design the website's frontend. Also, the collected data are stored in a MySQL database. Furthermore, since the models are trained on English corpus, we use Google translate API to identify the language of the claims and tweets' content that is written in languages other than English and convert them into English. Checkovid contains pre-trained machine learning and deep learning models and employs all proposed processes in this paper that enable users to:

1) Check the veracity of their claim automatically in paragraph-level using ten models.
2) Check the veracity of their claim in sentence-level that works by segmenting the claim into sentences and predicting the veracity of each sentence using the selected model.
3) Check the veracity of a tweet in both content-based and network-based by extracting features from the tweet's URL using Twitter API.
4) Check the similarity of their claim with other claims in our database.
5) Like or dislike the model decision in the sentence-level section for future work.
6) Access the collected datasets used in this paper that the research community can use.

### *4.2. Experimental Setting*

We implement all codes in python 3.7.2 and conduct our experiments on the Google Colab, a cloud platform that allows developers to create and train machine learning and deep learning models on CPUs, GPU, and TPU. We use Tensorflow 2.4.1 and scikit-learn 0.22.2.post1 version for implementing deep learning and machine learning models, respectively. Also, we use TfidfVectorizer in scikit-learn for vectorizing text and one_hot in TensorFlow to convert text into the numerical format. We use 60% of the data for the training set, 20% for the validation set, and 20% for the testing set to train and evaluate the models. Each deep learning model is trained on the training set for 10 epochs, and the loss on the validation set is used to pick the best epoch, and we evaluate the models with the test set. The models are optimized for the binary cross-entropy loss function using Adam optimizer with a batch size of 64. For text preprocessing, we use NLTK, and for tuning hyperparameters in the traditional machine learning model using the grid search method, we perform GridSearchCV provided by scikit-learn.

[4] http://checkovid19.com

## 4.3. *Metrics*

There are various metrics for evaluating the performance of a classification model. In the paper, we used Accuracy, Precision, Recall, and $F_1$ score.

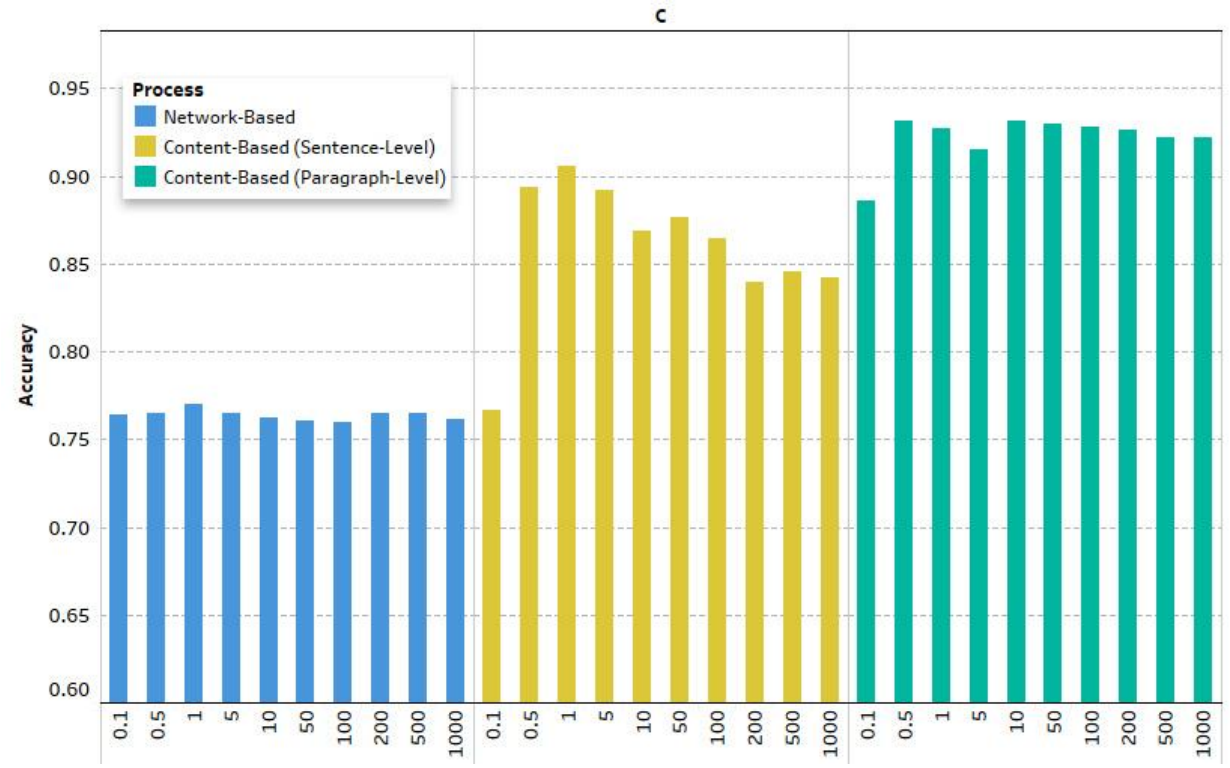

*(a) Logistic Regression*

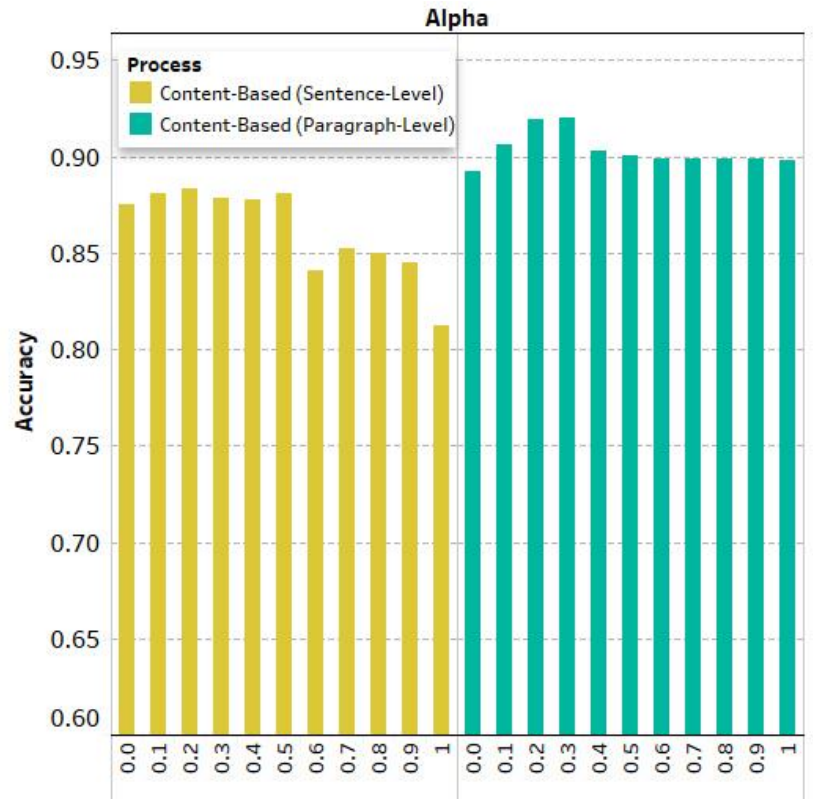

*(b) Naïve Bayes*

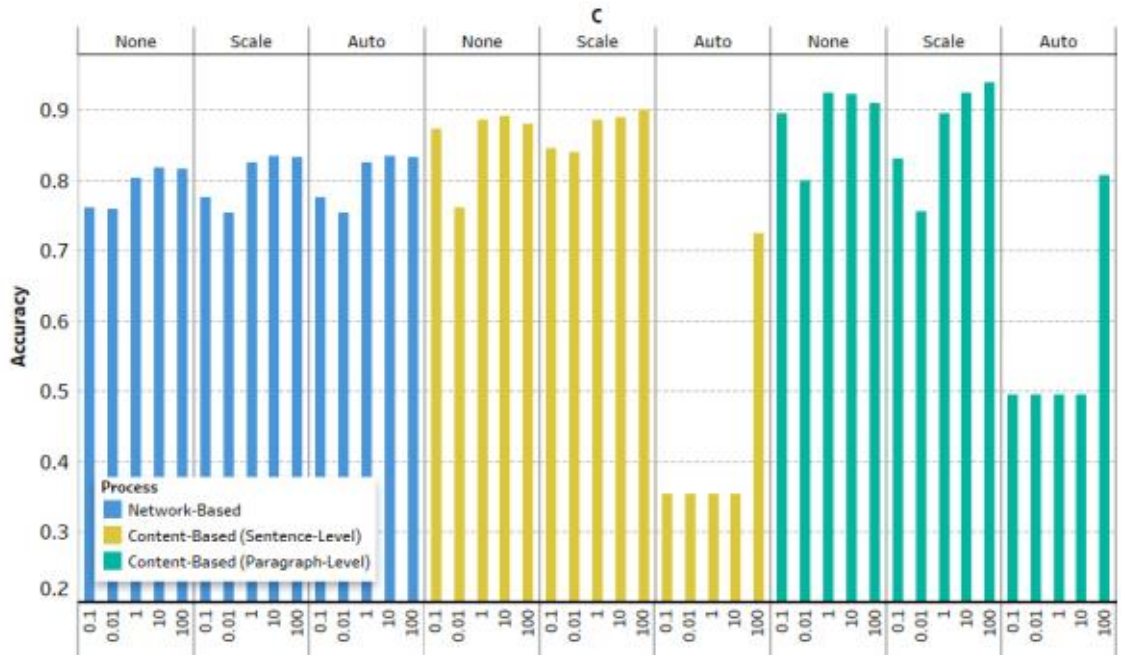

*(c) SVM*

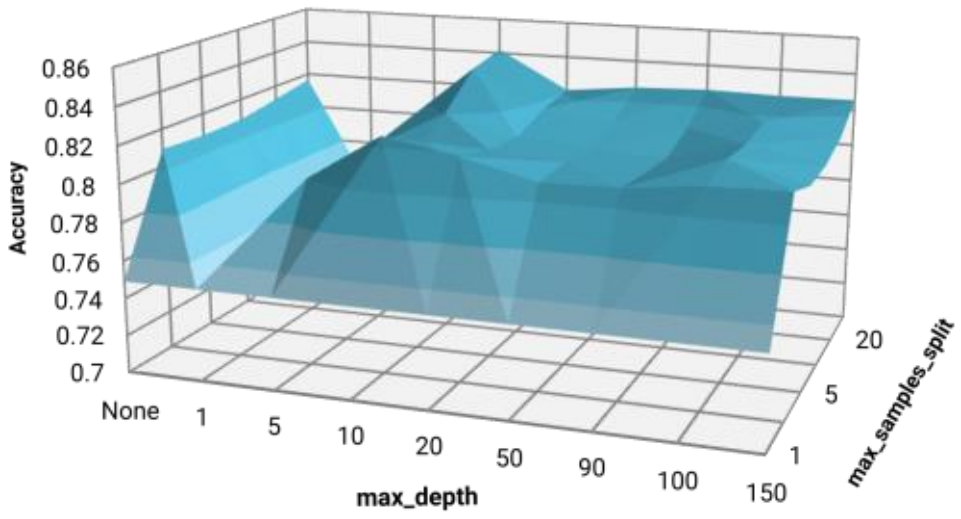

*(d) Decision Tree - Network-Based*

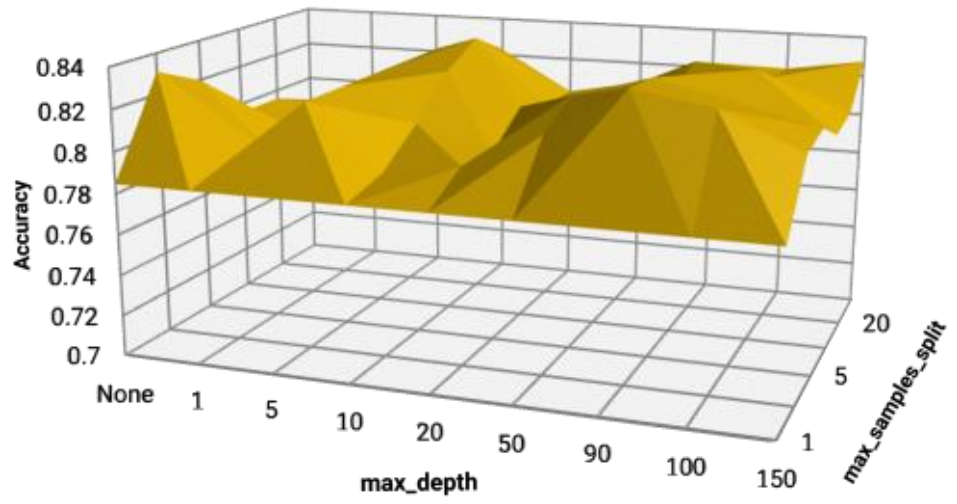

*(e) Decision Tree - Content-Based (Sentence-Level)*

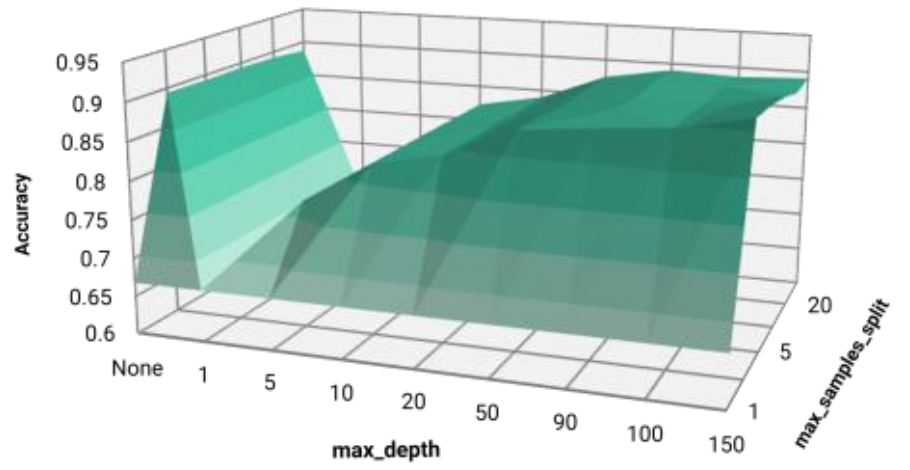

*(f) Decision Tree - Content-Based (Paragraph-Level)*

*Figure 14. The performance of the hyperparameters of the traditional machine learning models on the validation set*

### 4.3.1. *Hyperparameter Tuning*

To tune the hyperparameters of the traditional machine learning algorithms, we use the grid search method, which searches through a subset of hyperparameters of a model with the best result. The possible and the optimum values of the hyperparameters for each traditional machine learning model are shown in Table 3.

In both Logistic Regression and SVM models, we apply the L2 penalty to reduce overfitting, so we tune the value of the regularization parameter (C). Also, we perform the SVM model with linear kernel and RBF kernel. For the Naïve Bayes models, we tune the smoothing parameter (alpha) that handles the problem of zero probability. In the decision tree models, we analyze different parameters: functions to measures the quality of split (criterion), the maximum depth of the tree (max_depth), the maximum number of the features to consider for the best split (max_features), the minimum number of samples for splitting an internal node (min_samples_split), and the minimum number of the samples to be at a leaf node (min_samples_leaf). The performance of each parameter in the models on the validation set also has been illustrated in Figure 14.

*Table 3. Hyperparameter tuning performance using the Grid search method for traditional machine learning models*

| Models | Tuning Parameter | Optimal Parameter | | |
|---|---|---|---|---|
| | | Content-Based Paragraph-level | Content-Based Sentence-level | Network-Based |
| Logistic Regression | C=[0.1,0.5,1,5,10,50,100,200,500,1000] | C=0.5 | C=1 | C=500 |
| Naïve Bayes | alpha=[0,0.1,0.2,0.3,0.4,0.5,0.6,0.7,0.8,0.9,1] | alpha=0.3 | alpha=0.2 | No parameter |
| SVM | C=[0.01, 0.1, 1, 10,100]<br>kernel=['linear', 'rbf']<br>gamma=['scale', 'auto'] | C=100<br>kernel='rbf'<br>gamma='scale' | C=100<br>kernel='rbf'<br>gamma='scale' | C=10<br>kernel='rbf'<br>gamma='auto' |
| Decision Tree | criterion=['gini', 'entropy']<br>max_depth=[None,1,5,10,20,50,90,100,150]<br>max_features=[None, 'sqrt', 'auto', 'log2']<br>min_samples_split=[1, 2, 5, 10, 20, 40]<br>min_samples_leaf=[1, 2, 5, 10, 20] | criterion='entropy'<br>max_depth=150<br>max_features=None<br>min_samples_split=2<br>min_samples_leaf=1 | Criterion='entropy'<br>max_depth=90<br>max_features=None<br>min_samples_split=2<br>min_samples_leaf=1 | Criterion='entropy'<br>max_depth=10<br>max_features=None<br>min_samples_split=2<br>min_samples_leaf=20 |

Moreover, we tune different hyperparameters of the deep learning models, such as the number of hidden layers, the number of units, dropout rate. We evaluate the models' performance on the validation set and select models with the best architecture for each type of deep learning algorithm. The results of each model are shown in Figure 15.

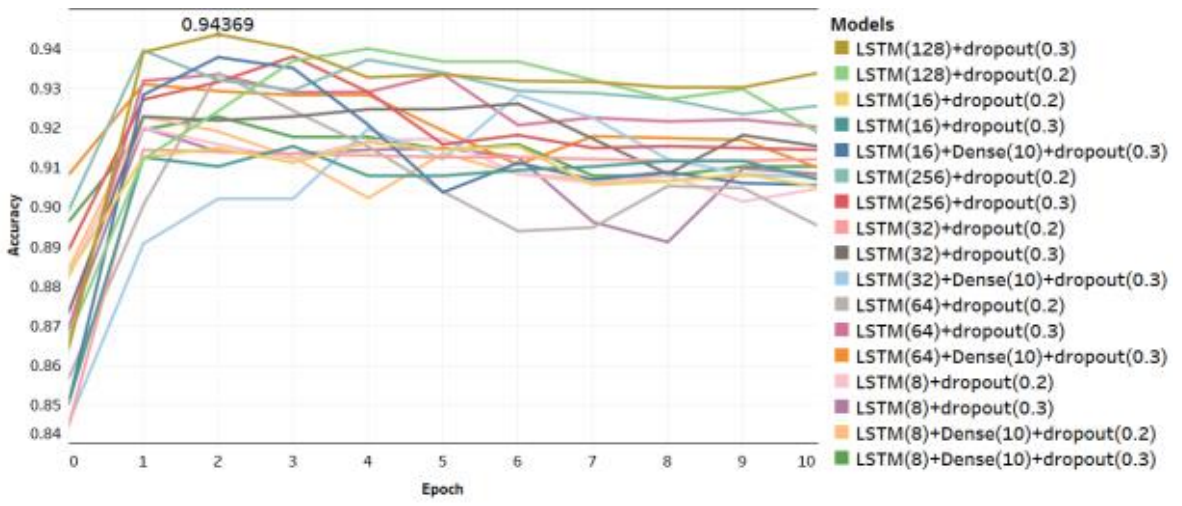

*(a) LSTM (text classification models)*

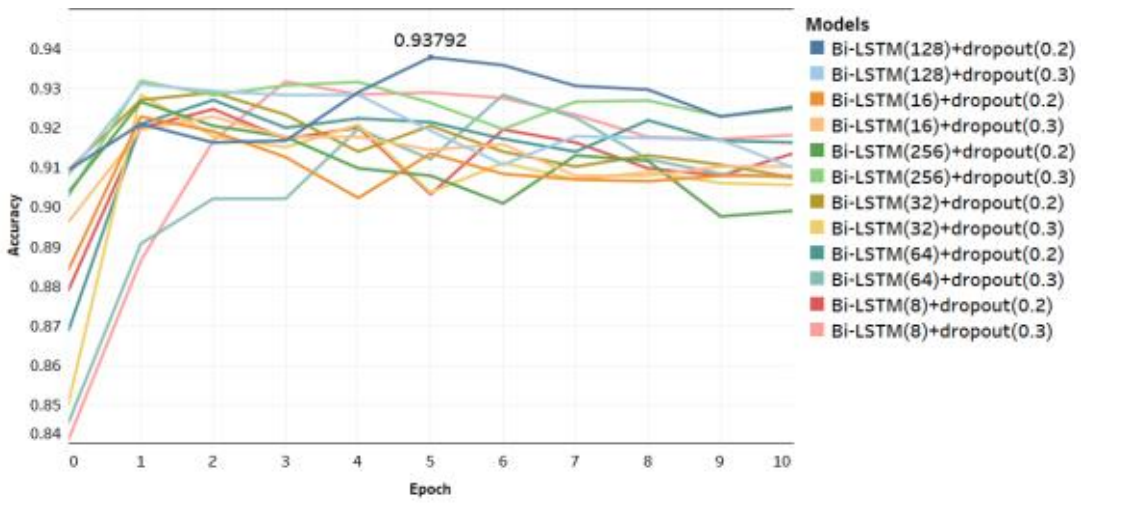

*(b) Bi-LSTM (text classification models)*

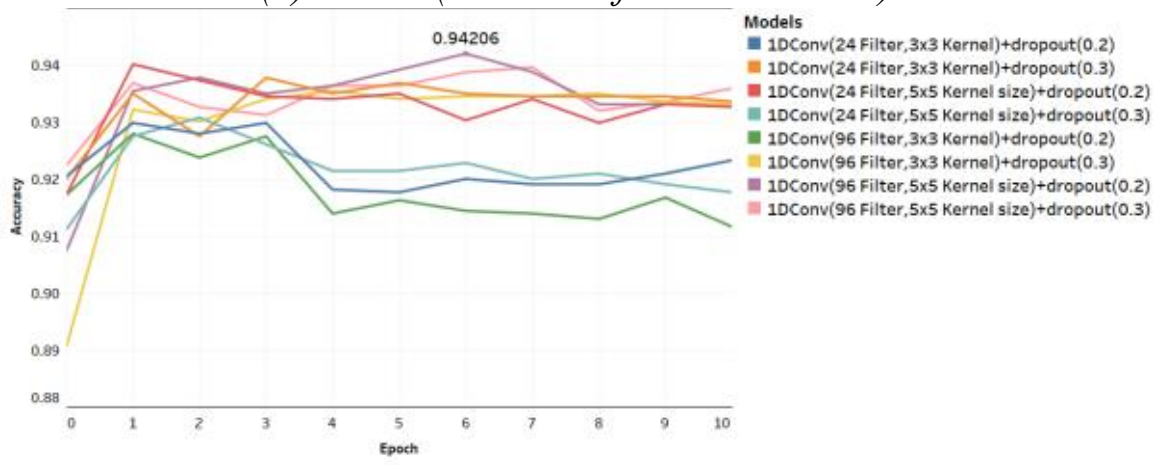

*(c) CNN (text classification models)*

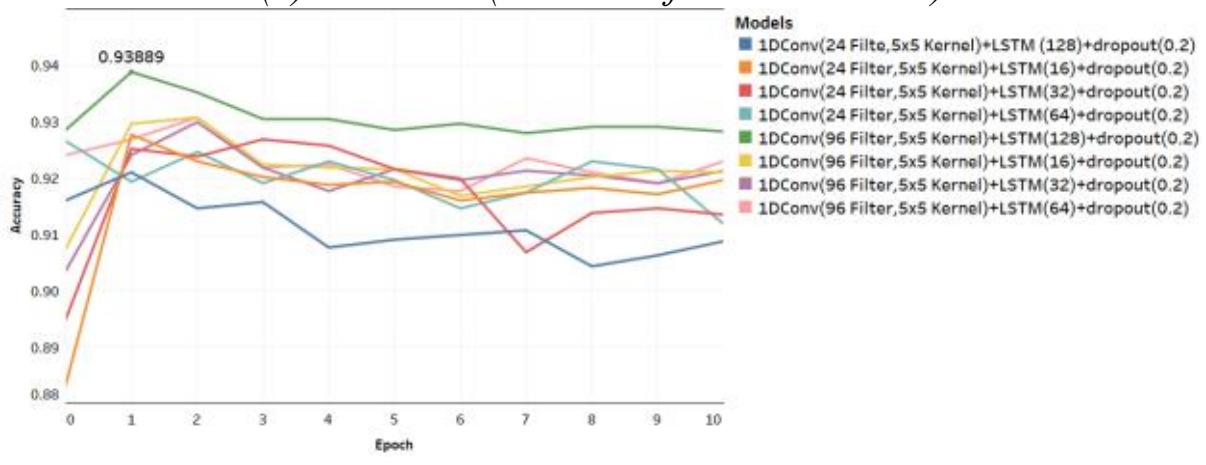

*(d) CNN+LSTM (text classification models)*

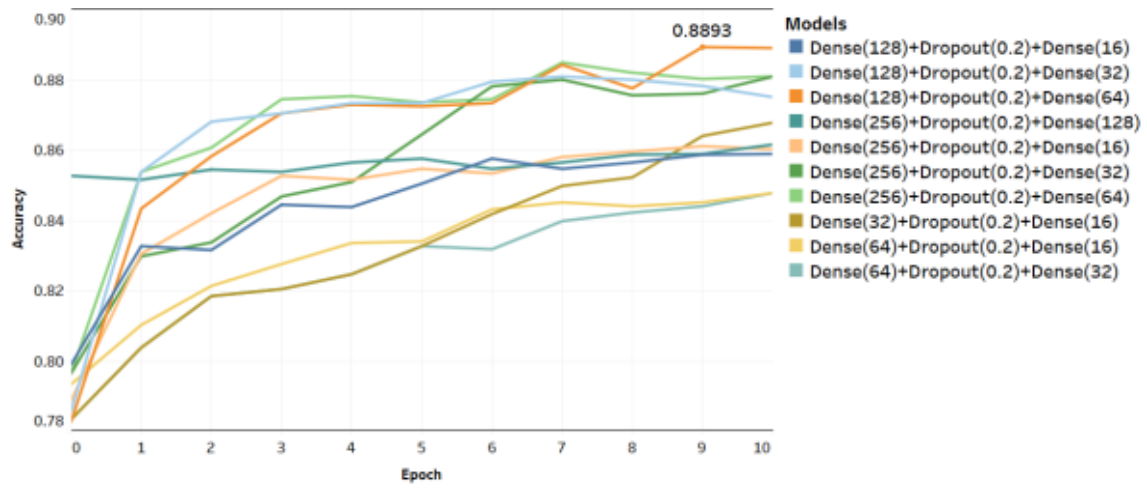

*(e) ANN (Network-Based)*

*Figure 15. The performance of different hyperparameters in deep learning models on the validation set*

- **Artificial Neural Network (Network-Based):**

In the input layer, there are 128 units in a dense layer with a ReLU activation function followed by a dropout layer with a rate of 0.2 and a dense layer with 64 units and a ReLU activation function. Finally, a one-unit dense layer is added to the architecture with sigmoid activation in the last layer.

- **LSTM (Content-Based):**

As the model's input, we pass the same-length encoded text into an embedding layer in the first layer. The following layers had a single LSTM layer with 128 units, one dropout layer with a 0.3 dropout rate, and a one-unit dense layer with a sigmoid activation function.

- **Bi-LSTM (Content-Based):**

This model has the same embedding layer as the previous model, followed by a bidirectional LSTM layer with 128 units, a dropout layer with a 0.2 dropout rate, and one dense layer with a single unit and a sigmoid activation function.

- **CNN (Content-Based):**

We develop a CNN model with an embedding layer as the first layer, followed by one 1D Conv layer with a ReLU activation function. The number of filters and the kernel size used in this model equals 96 and 5, respectively. The following layers have a 1D max-pooling layer and a dropout layer with a dropout rate of 0.2. Finally, in the last two layers, two dense layers are added to the architecture, one dense layer with 10 units and a ReLU activation function and a one-unit dense layer with sigmoid activation.

- **CNN+LSTM (Content-Based):**

This model is a combination of the LSTM and CNN model. After the embedding layer, just like other previous models, it has a 1D Conv layer with 96 filters followed by a max-pooling layer and a dropout layer with a 0.2 dropout rate. The output of this layer passed into the LSTM layer with 128 units and one dense layer with a sigmoid activation function.

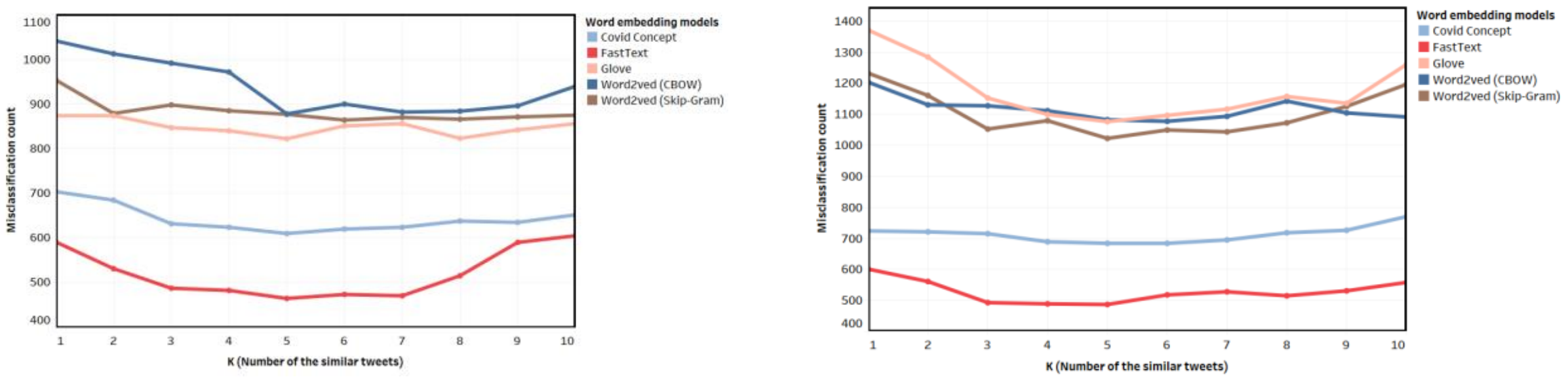

*(a) Cosine similarity* *(b) Euclidean Distance*

*Figure 16. Parameter tuning for similarity models*

In the similarity models, we evaluate models with different values of K to determine the best number of similar tweets to contribute to the classification. The misclassification errors of similarity models with the value of K from 1 to 10 are shown in Figure 16. Based on the results, the value of 1, which means assigning the label of the most similar tweet to the unlabeled tweet, has the worst result. The best result is obtained by the contribution of five similar texts for judging the veracity of a tweet (K=5).

### 4.3.2. *Performance of Network-Based models*

The results of the network-based models on the test set are reported in Table 4. We compare our model based on the mentioned metrics and the best performance among the network-based models obtained with the Artificial Neural Network (ANN) model by achieving an 88.68% $F_1$ score.

*Table 4. Result of Network-based classification on the test set*

| Models | | Accuracy | Precision | Recall | $F_1$ |
|---|---|---|---|---|---|
| Traditional Machine Learning | Logistic Regression | 0.7648 | 0.7612 | 0.7832 | 0.7720 |
| | Naive Bayes | 0.7342 | 0.7049 | 0.8208 | 0.7584 |
| | SVM | 0.8351 | 0.7993 | 0.9020 | 0.8476 |
| | Decision Tree | 0.8253 | 0.8259 | 0.8317 | 0.8287 |
| Ensemble Learning | Random Forest | 0.8724 | 0.8500 | 0.9095 | 0.8788 |
| | Stacking | 0.8484 | 0.8258 | 0.8894 | 0.8564 |
| Deep Learning | Artificial Neural Network | 0.8820 | 0.8652 | 0.9095 | 0.8868 |

### 4.3.3. *Performance of Content-Based models*

#### *4.3.3.1. Text classification models*

We developed different text classification models on two sets of datasets: *Dataset I* for paragraph-level and *Dataset II* for sentence-level. The best result in paragraph-level with a difference of 0.42% for $F_1$ score compared to the LSTM model achieved by the Stacking ensemble learning models with 95.18% for $F_1$ score. Also, we obtained the best result for sentence-level with the Stacking ensemble learning models with 90.25% for $F_1$ score (0.53% better than the LSTM model). The results of content-based models have shown in Table 5 and indicate better performance than the network-based process and better results of the ensemble-based model compared to deep learning models.

*Table 5. Result of text classification models in the Content-based process on the test set*

| | | Paragraph-Level | | | | Sentence-Level | | | |
|---|---|---|---|---|---|---|---|---|---|
| Models | | Accuracy | Precision | Recall | $F_1$ | Accuracy | Precision | Recall | $F_1$ |
| Traditional machine learning | Logistic Regression | 0.9361 | 0.9295 | 0.9449 | 0.9371 | 0.8942 | 0.8997 | 0.8870 | 0.8933 |
| | Naive Bayes | 0.9183 | 0.8808 | 0.9691 | 0.9228 | 0.8897 | 0.8872 | 0.8928 | 0.8899 |
| | SVM | 0.9327 | 0.9357 | 0.9305 | 0.9331 | 0.8905 | 0.8927 | 0.8875 | 0.8901 |
| | Decision Tree | 0.9144 | 0.9202 | 0.9090 | 0.9146 | 0.8377 | 0.8603 | 0.8060 | 0.8323 |
| Ensemble learning | Random forest | 0.9311 | 0.8943 | 0.9790 | 0.9347 | 0.8884 | 0.8994 | 0.8744 | 0.8867 |
| | Stacking | 0.9511 | 0.9482 | 0.9554 | 0.9518 | 0.9031 | 0.9071 | 0.8980 | 0.9025 |
| Deep learning | LSTM | 0.9469 | 0.9427 | 0.9526 | 0.9476 | 0.8979 | 0.9018 | 0.8928 | 0.8972 |
| | Bi-LSTM | 0.9361 | 0.9352 | 0.9382 | 0.9367 | 0.8958 | 0.8958 | 0.8875 | 0.8916 |
| | CNN | 0.9402 | 0.9259 | 0.9581 | 0.9417 | 0.8947 | 0.8887 | 0.8940 | 0.8914 |
| | CNN + LSTM | 0.9344 | 0.9462 | 0.9223 | 0.9341 | 0.8769 | 0.8688 | 0.8777 | 0.8732 |

#### *4.3.3.2. Similarity models*

We evaluate the performance of each word embedding model by splitting the data into a training and test set and classifying each tweet in the test set based on their similarity to tweets in the training set. The results are shown in Table 6 and indicate the improvement in the classification results of our novel models compared to the models in the network-based process. We obtain the best performance of 90.26% using the cosine similarity and fastText word embedding. Since two similar texts can be far apart in space but could still have a small angle between them, models with the cosine similarity have better results than models with Euclidean distance. Moreover, the fastText model, which uses n-grams, has significant improvement than Word2vec.

*Table 6. Result of Similarity models in Content-based process on the test set*

| | Cosine Similarity | | | | Euclidean Distance | | | |
|---|---|---|---|---|---|---|---|---|
| Word Embeddings | Accuracy | Precision | Recall | $F_1$ | Accuracy | Precision | Recall | $F_1$ |
| Word2vec (CBOW) | 0.7726 | 0.7622 | 0.8002 | 0.7807 | 0.7566 | 0.7513 | 0.7792 | 0.7650 |
| Word2vec (Skip-Gram) | 0.7951 | 0.7723 | 0.8434 | 0.8063 | 0.7733 | 0.7641 | 0.8015 | 0.7824 |
| Glove | 0.7870 | 0.7468 | 0.8757 | 0.8061 | 0.7200 | 0.6909 | 0.8129 | 0.7469 |
| FastText | 0.8971 | 0.8695 | 0.9383 | 0.9026 | 0.8918 | 0.8866 | 0.9025 | 0.8945 |
| COVID-19 Concept | 0.8404 | 0.8455 | 0.8395 | 0.8425 | 0.8310 | 0.7959 | 0.8955 | 0.8428 |

Also, the COVID-19 Concept Embedding, which trained on the dataset containing scholarly articles and claims about COVID-19 and related coronaviruses, achieved a better classification result than Glove and Word2vec embeddings.

#### *4.3.3.3. Constraint@AAAI2021 shared task*

Besides testing the models with our dataset, we apply our content-based models on the Constraint@AAAI2021 COVID-19 fake news detection dataset (Patwa et al., 2020). This dataset was provided for the shared task containing 10700 REAL and FAKE English posts and articles that have been manually annotated and collected from various social media and fact-checking websites. Based on the results shown in Table 7, we achieved a maximum $F_1$ score of 94.38% using the LSTM model over the baseline $F_1$ score of 93.46% among text classification models. We obtained a 90.75% $F_1$ score among similarity models by the fastText word embedding with the cosine similarity metric.

*Table 7. Result of text classification models and similarity models on the Constraint@AAAI2021 Dataset*

| Models | | | accuracy | Precision | Recall | $F_1$ |
|---|---|---|---|---|---|---|
| Text classification models | Traditional machine learning | Logistic Regression | 0.9303 | 0.9442 | 0.9194 | 0.9316 |
| | | Naive Bayes | 0.9186 | 0.9338 | 0.9067 | 0.9201 |
| | | SVM | 0.9214 | 0.9317 | 0.9149 | 0.9232 |
| | | Decision Tree | 0.8546 | 0.8642 | 0.8524 | 0.8583 |
| | Ensemble learning | Random forest | 0.9214 | 0.9302 | 0.9167 | 0.9234 |
| | | Stacking | 0.9406 | 0.9527 | 0.9312 | 0.9418 |
| | Deep learning | LSTM | 0.9420 | 0.9438 | 0.9438 | 0.9438 |
| | | Bi-LSTM | 0.9308 | 0.9353 | 0.9303 | 0.9328 |
| | | CNN | 0.9416 | 0.9406 | 0.9438 | 0.9422 |
| | | CNN + LSTM | 0.9299 | 0.9290 | 0.9357 | 0.9323 |
| Similarity models | | Word2vec (CBOW) | 0.7581 | 0.7871 | 0.7560 | 0.7712 |
| | | Word2vec (Skip-Gram) | 0.8078 | 0.8174 | 0.8288 | 0.8231 |
| | | Glove | 0.8041 | 0.7720 | 0.9036 | 0.8326 |
| | | FastText | 0.9001 | 0.9072 | 0.9078 | 0.9075 |
| | | COVID-19 Concept | 0.8523 | 0.8168 | 0.9362 | 0.8724 |

### 4.3.4. *Discussion*

Based on the results obtained in the previous section, shown in Figure 18, the content-based models perform better than network-based models to classify misinformation. The best results were obtained from the stacking ensemble model in terms of models with a slight difference from the LSTM model in paragraph-level text classification models. Also, the result of the similarity models indicates a large number of similar tweets written in the context of the COVID-19 disease. The similarity model using the cosine similarity and the fastText embedding outperformed the network-based models. Furthermore, since predicting a false and misinformative claim as a valid claim poses a greater danger than expecting a true claim as a false claim, text classification models compared to other models better distinguish false-negative (Figure 17).

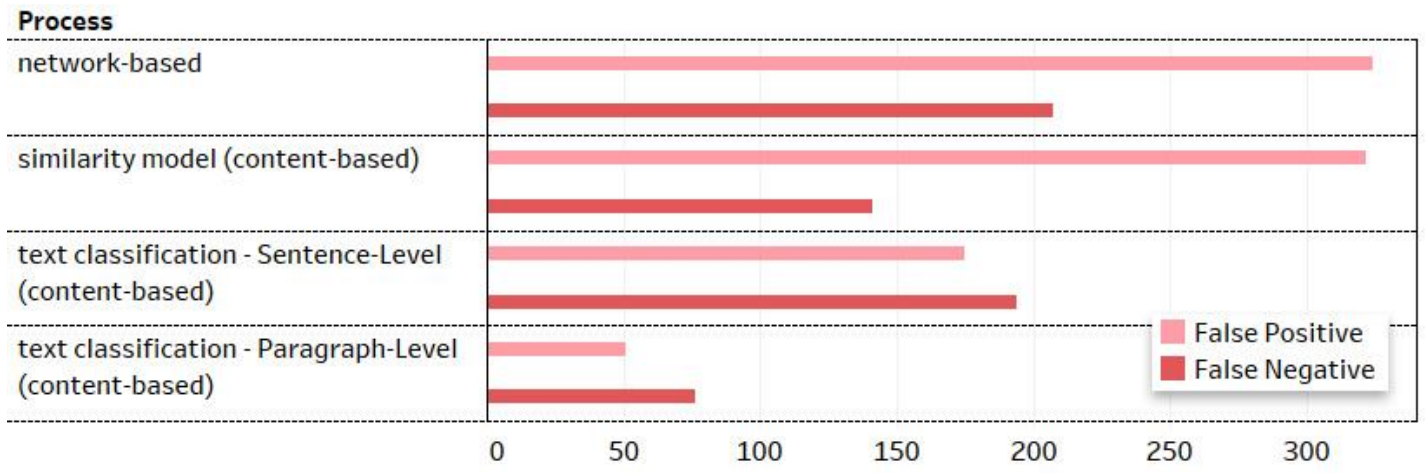

*Figure 17. Number of False Negative and False Positive in the best model of each process on the test set*

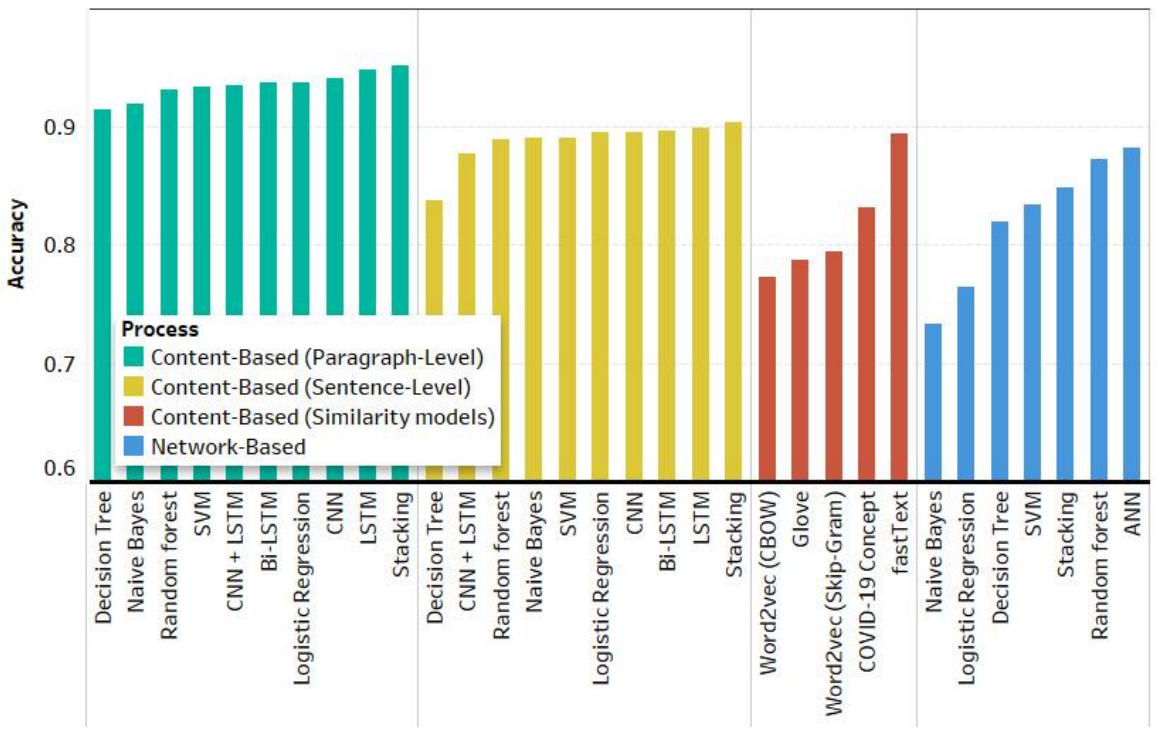

*Figure 18. The results of models on the test set*

## 5. Conclusion

With the emergence of a crisis such as the COVID-19 pandemic, a range of new challenges come to light. During the COVID-19 pandemic, one of these challenges was the misinformative claims that proliferated all over the social media platforms in a short amount of time. Therefore, it is exacerbated by the global scale of the emergency. The main contribution of this work is as follows. This paper gathered a COVID-19 related misinformation dataset on Twitter and constructed a dataset containing sentences divided into informative and misinformative. Also, we proposed a misinformation detection system that includes two processes called network-based and content-based applied to the collected datasets. Each process entails NLP techniques and machine learning models and addresses the issue of distinguishing the veracity of tweets in the domain of COVID-19 from various aspects. We discovered that detecting whether a tweet is informative or misinformative, based solely on the network and user characteristics, might not be the best idea.

Moreover, we showed that our novel similarity models that can detect misinformation based on the similarity of texts outperform the network-based models due to many similar tweets. Furthermore, we developed the text classification models in paragraph-level and sentence-level that are trained on the content of the tweets. The results revealed that the text classification models had a better performance than the other models and showed that the ensemble learning models outperform the deep learning models with a slight difference. Overall, we obtained the best performance of 95.18% $F_1$ score with the stacking ensemble-learning text classification model in the content-based process. Finally, we developed a fact-checking website called Checkovid that utilizes each process to detect misinformation about COVID-19.

This work can help detect misinformation about any future global health crisis and other not health-related domains by changing the dataset and the source of information. Also, it can be further extended by utilizing all of the processes in a hybrid system, and we can use the dataset that will be obtained from users' comments (like or dislike) on the website to improve our work. Furthermore, based on the results of the similarity-based models in classification, which are performed on the content of the tweets, we can use network-based features to classify misinformation based on the similarity of these features.